\documentclass[runningheads]{llncs}
\usepackage[T1]{fontenc}

\usepackage{graphicx,verbatim}

\usepackage{amssymb}
\usepackage{lipsum}
\usepackage{xstring}
\usepackage{xcolor}

\usepackage{booktabs}   
\usepackage{multirow}   
\usepackage{caption}    
\usepackage[misc]{ifsym} 
\usepackage{float} 

\begin{document}
\title{PromptGate: Client-Adaptive Vision–Language Gating for Open-Set Federated Active Learning}
\titlerunning{PromptGate}

\author{
Adea Nesturi\textsuperscript{$\star$} \and
David Dueñas Gaviria\textsuperscript{$\star$} \and
Jiajun Zeng \and
Shadi Albarqouni\textsuperscript{\Letter}
}

\authorrunning{Nesturi and D. Gaviria et al.}
\institute{
University of Bonn, University Hospital Bonn, Clinic for Diagnostic and Interventional Radiology, 53127 Bonn, Germany\\
\email{\{Adea.Nesturi,David.Duenas-Gaviria,Shadi.Albarqouni\}@ukbonn.de}
}

\renewcommand\thefootnote{}
\footnotetext{\textsuperscript{$\star$} These authors equally contributed to this work.} 
\footnotetext{\textsuperscript{\Letter} Corresponding author: Shadi Albarqouni (Shadi.Albarqouni@ukbonn.de)}
  
\maketitle             
\begin{abstract}
Deploying medical AI across resource-constrained institutions demands data-efficient learning pipelines that respect patient privacy. Federated Learning (FL) enables collaborative medical AI without centralising data, yet real-world clinical pools are inherently open-set, containing out-of-distribution (OOD) noise such as imaging artifacts and wrong modalities. Standard Active Learning (AL) query strategies mistake this noise for informative samples, wasting scarce annotation budgets. We propose \textbf{PromptGate}, a dynamic VLM-gated framework for Open-Set Federated AL that purifies unlabeled pools before querying. PromptGate introduces a federated Class-Specific Context Optimization: lightweight, learnable prompt vectors that adapt a frozen BiomedCLIP backbone to local clinical domains and aggregate globally via FedAvg---without sharing patient data. As new annotations arrive, prompts progressively sharpen the ID/OOD boundary, turning the VLM into a dynamic gatekeeper that is \emph{strategy-agnostic}: a plug-and-play pre-selection module enhancing any downstream AL strategy. Experiments on distributed dermatology and breast imaging benchmarks show that while static VLM prompting degrades to ${\sim}50\%$ ID purity, PromptGate maintains $>$95\% purity with 98\% OOD recall. 

\keywords{Active Learning  \and Federated Learning \and Out-of-distribution.}

\end{abstract}
\section{Introduction}
\label{sec:intro}
Deep learning has achieved impressive performance in medical image analysis, but still depends on large, carefully curated labeled datasets, which are scarce in routine clinical workflows. Active learning (AL) mitigates this by iteratively selecting informative unlabeled samples for annotation~\cite{settles2009active}, and is a natural candidate to reduce pathologists' labeling effort. Most classical AL, however, assumes a \emph{closed-set} scenario in which all pool images belong to the target classes. In contrast, hospital archives contain a heterogeneous mix of normal tissue, artifacts, benign findings, and unrelated pathologies, so naïve AL often spends much of its budget on out-of-distribution (OOD) or non-target samples.

Open-set active learning (OAL) addresses this by separating in-distribution (ID) from OOD samples to maximize query \emph{purity}. Recent OAL methods combine feature-based candidate sets with uncertainty sampling~\cite{qu2023openal}, margin-based detectors~\cite{Tang2024OSALNDOA}, or joint OOD filtering and class discovery~\cite{schmidt2025joint}. OpenPath~\cite{zhong2025openpath} shows that pre-trained vision--language models (VLMs) can warm up open-set AL via target/non-target text prompts in a centralized setting. These works, however, assume a single pooled dataset and treat OOD as a fixed global notion, which does not reflect real multi-institutional data.

In practice, data are fragmented across sites with different scanners and staining protocols. Federated learning (FL)~\cite{mcmahan2017communication} trains models without sharing raw data, and its combination with AL has produced federated AL (FAL) methods that reduce annotation cost under data-governance constraints~\cite{huang2023rethinking,Chen2024FEAL}. Extending this to \emph{open-set federated active learning} (OS-FAL) is natural, yet existing approaches rely on task-specific representations and treat OOD geometrically in feature space, without leveraging VLM semantic priors.

Prompt learning~\cite{zhou2022coop} adapts frozen VLM backbones (e.g., BiomedCLIP~\cite{biomedclip}) by optimizing continuous context tokens, capturing task-specific semantics from a few labels. We take an orthogonal approach to OpenPath: rather than using VLMs to define an acquisition rule, we use them as a \emph{client-adaptive filtering front-end} that can precede any OS-FAL strategy.

Concretely, we propose \emph{PromptGate}, a lightweight dynamic VLM-gated module for OS-FAL. Prompts are split into \emph{global} tokens (shared via FedAvg) and \emph{local} tokens (personalized per site), enabling a frozen VLM to pseudo-label the unlabeled pool and form a high-confidence ID candidate set passed to any downstream acquisition rule (e.g., Random, Entropy). Oracle labels then update both prompt components in a CoOp-style fashion~\cite{zhou2022coop}, progressively aligning VLM semantics with each site's notion of target vs.\ non-target. Our \textbf{contributions} are threefold: we introduce PromptGate, the first learnable-prompt VLM module for OS-FAL, with global/local prompt decomposition to capture heterogeneous OOD behavior; we propose a VLM-based gating mechanism that forms a high-purity candidate pool for any downstream acquisition rule; and we demonstrate plug-and-play improvements in query purity and label efficiency across multiple AL strategies and two federated medical imaging benchmarks.

\section{Method}
\label{sec:method}

\begin{figure}[t!]
    \centering
    \label{fig:teaser}
    \includegraphics[width=1.0\textwidth]{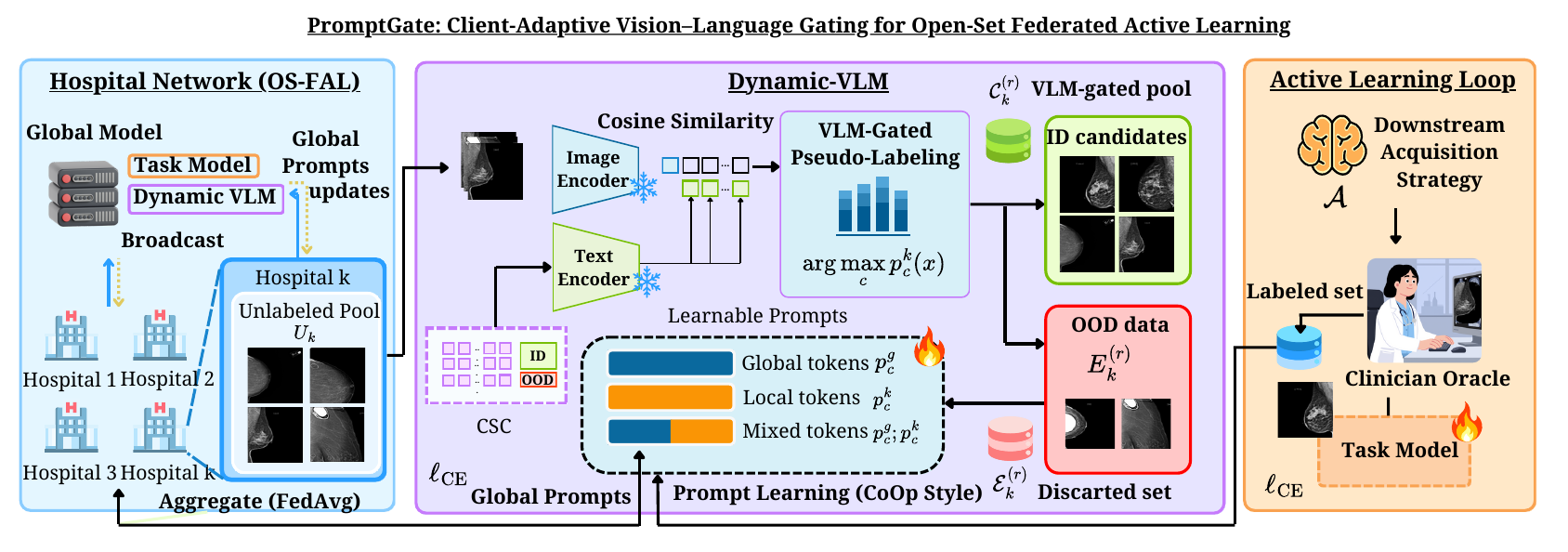}
    \caption{Overview of PromptGate for OS-FAL.}
    \label{fig:framework}
\end{figure}

We now describe \emph{PromptGate} (Fig. \ref{fig:framework}), our client-adaptive vision--language gating module for OS-FAL.
We consider a $C$-class task with a central server and $K$ clients.
Client $k$ holds a labeled set $L_k$ and an unlabeled pool $U_k = U_{k,\mathrm{ID}} \cup U_{k,\mathrm{OOD}}$, where $U_{k,\mathrm{ID}}$ are target images and $U_{k,\mathrm{OOD}}$ non-target ones (artifacts, background, other pathologies).
At each AL round $r$, the server broadcasts model and prompt parameters; each client uses a VLM to construct a \emph{gated} ID candidate pool from $U_k^{(r)}$, applies any acquisition strategy $\mathcal{A}$, and updates its local model and prompts with the newly acquired labels.

\subsection{VLM with Global and Local Learnable Prompts}

PromptGate assumes a pre-trained VLM with frozen image encoder $E_{\mathrm{img}}$ and text encoder $E_{\mathrm{text}}$.
For an image $x$, we obtain $z(x) = E_{\mathrm{img}}(x) \in \mathbb{R}^D$.
Following CoOp~\cite{zhou2022coop}, we model prompts as continuous context tokens optimised in the text embedding space while keeping VLM weights fixed.
We employ \textbf{Class-Specific Context (CSC)} optimization, assigning each ID class $c \in \{1,\dots,C\}$ and a single non-target label ``OOD'' unique learnable tokens \emph{factored} into:
(i) global tokens $p^g_c \in \mathbb{R}^{d_p \times D}$, aggregated across clients to capture a shared semantic prior; and
(ii) client-specific tokens $p^k_c \in \mathbb{R}^{d_p \times D}$, private to client $k$ to adapt to local data heterogeneity.
These matrices are concatenated to form a single context $[p^g_c; p^k_c]$.
Given a class-specific template $T_c$, we obtain the prompted text embedding
\[
    t^k_c = E_{\mathrm{text}}\!\big([p^g_c; p^k_c], T_c \big) \in \mathbb{R}^D,
    \qquad c \in \{1,\dots,C,\mathrm{OOD}\}.
\]

\subsection{VLM-Gated Pseudo-Labeling and Acquisition}
For each $x \in U_k^{(r)}$, client $k$ computes cosine similarities
$s^k_c(x) = {z(x)^\top t^k_c}\big/\big({\|z(x)\| \, \|t^k_c\|}\big)$
and converts them into pseudo-label probabilities via temperature-scaled softmax:
$p^k_c(x) \propto \exp\!\big(s^k_c(x)/\tau_{\mathrm{VLM}}\big)$, with pseudo-label $\hat{y}^k(x) = \arg\max_{c}\, p^k_c(x)$.

We turn these scores into a pseudo-label distribution via a temperature-scaled softmax
$p^k_c(x) \propto \exp(s^k_c(x)/\tau_{\mathrm{VLM}})$, and define:
\[
    \hat{y}^k(x) = \arg\max_{c} p^k_c(x), 
    \qquad
    p^k_{\max}(x) = \max_{c} p^k_c(x).
\]
PromptGate then filters $U_k^{(r)}$ into an \textbf{ID-candidate pool (VLM-gated)}:
\[
    \mathcal{C}_k^{(r)} =
    \big\{ x \in U_k^{(r)} \; \big| \;
    \hat{y}^k(x) \in \{1,\dots,C\} \big\},
\]
retaining samples whose most probable VLM class is an ID category.
The discarded pool $\mathcal{E}_k^{(r)} = U_k^{(r)} \setminus \mathcal{C}_k^{(r)}$ serves as quality control; empirically, ID leakage into $\mathcal{E}_k^{(r)}$ is negligible.
Any AL strategy $\mathcal{A}$ then selects queries from the gated pool:
$Q^{(r)}_{k} = \mathcal{A}\big(\mathcal{C}_k^{(r)}, L_k^{(r)}, \theta_k^{(r)}\big)$,
where $\theta_k^{(r)}$ is the current client model.
PromptGate thus acts as a plug-in \emph{pre-selection gate} in front of any OS-FAL rule.

\paragraph{\textbf{Acquisition on the Gated Pool.}}

The VLM-gated pool $\mathcal{C}_k^{(r)}$ is passed to any AL acquisition strategy $\mathcal{A}$ (e.g., Random, Entropy or FEAL):
\[
    Q^{(r)}_{k} = 
    \mathcal{A}\big(\mathcal{C}_k^{(r)}, L_k^{(r)}, \theta_k^{(r)}\big),
\]
where $\theta_k^{(r)}$ denotes the current client model.
PromptGate acts purely as a plug-in \emph{pre-selection gate} in front of the chosen OS-FAL rule.

\subsection{Prompt Learning and Federated Aggregation}

After querying, the oracle provides true labels for all $x \in Q^{(r)}_{k}$: either an ID class $y \in \{1,\dots,C\}$ or a coarse non-target label.
Let $\phi^g = \{p^g_c\}_c$ and $\phi^k = \{p^k_c\}_c$ denote global and client-specific tokens.
We define a prompt loss:
\[    \mathcal{L}^{(r)}_{\mathrm{prompt},k}(\phi^g,\phi^k)
    =
    \frac{1}{|Q^{(r)}_{k}|}
    \sum_{x \in Q^{(r)}_{k}} \ell_{\mathrm{CE}}\big(p^k(x), y(x)\big),
\]
where $\ell_{\mathrm{CE}}$ is cross-entropy between VLM class probabilities $p^k(x)$ and the true label $y(x)$.
Each client performs gradient steps on $\phi^g,\phi^k$, then sends only the $\phi^g$ update to the server, which aggregates updates via FedAvg and broadcasts $\phi^{g,(r+1)}$.
Local tokens $\phi^{k,(r+1)}$ remain private.
Over rounds, global prompts capture a federated semantic prior for ID vs.\ non-target, while local prompts specialise to each site's OOD behaviour, progressively improving the ID candidate pool for any downstream OS-FAL rule.
\section{Experimental Design and Results}
\label{sec:exp}

\paragraph{Baselines.} We evaluate four experimental configurations: (i)~\textbf{Coldstart}, the standard federated AL baseline where all unlabeled samples (both ID and OOD) are available for querying without any filtering; (ii)~\textbf{UpperB}, an oracle baseline where only in-distribution samples exist in the unlabeled pool, serving as an upper bound on AL performance; (iii)~\textbf{Baseline} (VLM-Static), where a zero-shot BiomedCLIP model partitions the unlabeled pool into a Gated Pool (predicted ID) and an Exploration Pool (predicted OOD) at R=0, with the gate remaining fixed throughout all subsequent rounds; and (iv)~\textbf{Ours} (PromptGate), where the VLM gate is refined via federated CoOp prompt tuning after each AL round, with three configurations: Mixed (8G-L8), our default setting that combines global and local prompts; Global (16G), which relies solely on globally shared prompts; and Local (16L), which uses only client-specific prompts; allowing the ID/OOD separation to adapt as new labeled data is acquired.

\paragraph{Datasets.}

\textbf{FedISIC}~\cite{chen2024think}\cite{ogier2022flamby}: A public multi-center skin lesion benchmark with four sites (9,930; 3,163; 2,691; 1,807 samples) spanning eight classes. We simulate the OS-FAL scenario by injecting OOD samples (50\% ratio relative to the local unlabeled ID pool) from DDI~\cite{daneshjou2022disparities} and Fitzpatrick17k~\cite{groh2021evaluating}, capturing clinical shifts. Training data spans four clients with heterogeneous OOD prevalence (32.2\%; 35.4\%; 35.7\%; 32.5\%). The test set evaluates generalization across four centers with naturally varied OOD distributions: Center 0 (2,840 samples; 12.6\% OOD), Center 1 (900; 12.1\%), Center 2 (772; 13.0\%), and Center 3 (522; 13.4\%).
\textbf{FedEMBED}: A breast density classification benchmark from the EMBED dataset~\cite{jeong2023emory} distributed across 222,700 training and 65,891 test samples~\cite{roschewitz2409robust}. The data spans four primary scanners—Selenia Dimensions (198,726 train / 58,925 test), Senograph 2000D ADS (10,043 / 2,810), Lorad Selenia (7,749 / 2,421), and Clearview CSm (6,182 / 1,735). To assess real-world OOD, we consider as OOD the clinical artifacts annotated by~\cite{schueppert2025radio}, organically present in the data. The total OOD prevalence varies heterogeneously across test domains, driven by distinct artifact profiles: Selenia Dimensions (13.15\% total; predominantly 8.49\% circles), Senograph 2000D ADS (11.03\% total; predominantly 5.90\% implants), Lorad Selenia (12.38\% total; split between 5.95\% circles and 4.10\% implants), and Clearview CSm (4.78\% total; 4.60\% circles).

\paragraph{Evaluation Metrics.}
We report Balanced Multiclass Accuracy (BMA) on the test sets, Query Precision (QP; fraction of true ID samples in each query round), Accumulated Query Recall (AQR; cumulative fraction of the retrieved ID), and ID pool Purity (fraction of true ID samples in the ID-candidate pool).

\paragraph{Implementation details.}
All closed-set AL (Random, Entropy~\cite{shannon1948mathematical}, FAL (FEAL~\cite{Chen2024FEAL}) including OAL (PAL~\cite{yang2023not}, LfOSA~\cite{ning2022active}) are introduced to OS-FAL using their default configurations. Local training follows standard FedAvg (15 epochs, batch 32, lr 0.0005) across three seeds. We employ frozen BiomedCLIP~\cite{biomedclip} as the VLM backbone. Prompt tuning uses CSC CoOp with 8 global + 8 local learnable vectors per class (16 total) in the mixed configuration, trained via SGD (lr 0.002, momentum 0.9, weight decay 5e-4) for 15 epochs on a 128-shot subset per client. The AL budget is 500 samples/round (R=5 for FedISIC, R=10 for FedEMBED). Downstream classifiers are EfficientNet-B0 for FedISIC~\cite{chen2024think} and a linear probe for FedEMBED~\cite{germani2025bias}. For OpenPath*~\cite{zhong2025openpath}, we use its k-means strategy with a single static OOD prompt and no self-training, ensuring fair comparison. Base prompts follow dataset conventions: ``A dermoscopy image of [CLASS]'' / ``Unknown or artifact'' (FedISIC) and ``A mammogram showing [CLASS] breast density'' / ``Unknown or artifact'' (FedEMBED).

\subsection{PromptGate vs. Static and Coldstart}

\begin{table*}[t!]
    \centering
    \caption{Comparison of VLM Filtering Strategies. Average scores of ID Purity and BMA over three seeds on the last AL round are reported. \textbf{Bold} indicates best purity, and \underline{Underline} indicates best BMA within each dataset.}
    \label{tab:combined_purity_bma}
    \resizebox{\textwidth}{!}{%
    \begin{tabular}{lccccccc}
    \toprule
    \textbf{Strategies} & \textbf{Random} & \textbf{Entropy} & \textbf{FEAL} & \textbf{PAL} & \textbf{LfOSA} & \textbf{OpenPath*} & \textbf{Avg} \\
    \midrule
    \multicolumn{8}{c}{\textbf{FedISIC} (R = 5)} \\
    \midrule
    Coldstart & 61.7 (60.2) & 59.1 (62.0) & 45.9 (57.6) & 61.1 (50.5) & 52.1 (55.7) & 84.1 (60.5) & 60.7 (57.8) \\
    Baseline & 60.9 (59.5) & 72.9 (61.5) & 78.7 (58.0) & 73.1 (49.7) & 76.1 (54.1) & 83.8 (61.0) & 74.3 (57.3) \\
    Ours (Mixed) & 97.4 (64.4) & 98.9 (\underline{66.1}) & 98.7 (57.2) & \textbf{98.7} (50.4) & 98.6 (54.9) & 86.5 (63.5) & 96.5 (59.4) \\
    Ours (Global) & 98.1 (59.8) & 99.0 (63.7) & \textbf{98.9} (58.5) & 98.7 (\underline{53.1}) & 99.0 (58.3) & \textbf{86.6} (\underline{64.4}) & 96.7 (59.6) \\
    Ours (Local) & \textbf{98.2} (\underline{65.3}) & \textbf{99.5} (64.2) & 98.8 (\underline{60.1}) & 98.7 (50.2) & \textbf{99.2} (\underline{60.0}) & 86.5 (62.2) & \textbf{96.8} (\underline{60.3}) \\
    UpperB. & 100.0 (63.2) & 100.0 (67.2) & 100.0 (53.5) & 100.0 (62.9) & 100.0 (64.3) & 100.0 (60.2) & 100.0 (61.9) \\
    \midrule
    \multicolumn{8}{c}{\textbf{FedEMBED} (R = 10)} \\
    \midrule
    Coldstart & 89.6 (60.0) & 80.9 (59.2) & 82.6 (42.5) & 87.5 (58.0) & 91.4 (58.9) & 95.1 (60.8) & 87.9 (56.6) \\
    Baseline & 89.9 (59.9) & 80.2 (59.2) & 82.8 (42.5) & 87.7 (57.7) & 92.4 (58.7) & 95.0 (60.5) & 88.0 (56.4) \\
    Ours (Mixed) & \textbf{93.1} (\underline{60.6}) & 85.7 (59.4) & 87.0 (\underline{43.1}) & 88.8 (58.1) & 92.6 (\underline{59.0}) & \textbf{96.0} (\underline{60.6}) & 90.5 (\underline{56.8}) \\
    Ours (Global) & 92.6 (60.5) & 85.2 (59.3) & 86.8 (42.5) & 88.5 (57.6) & \textbf{92.7} (58.8) & 95.9 (60.6) & 90.3 (56.6) \\
    Ours (Local) & \textbf{93.1} (60.2) & \textbf{85.8} (\underline{59.5}) & \textbf{87.4} (42.9) & \textbf{89.2} (\underline{58.2}) & 92.7 (58.6) & 95.8 (60.2) & \textbf{90.7} (56.6) \\
    UpperB. & 100.0 (60.8) & 100.0 (59.9) & 100.0 (43.4) & 100.0 (57.5) & 100.0 (58.6) & 100.0 (60.2) & 100.0 (56.7) \\
    \bottomrule
    \end{tabular}%
    }
\end{table*}

\begin{figure*}[t!]
    \centering
    \includegraphics[width=0.7\textwidth]{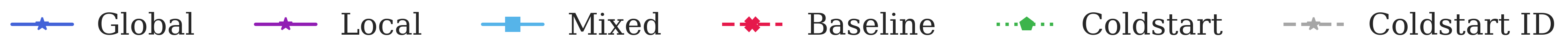}\\ 
    \includegraphics[width=0.7\textwidth]{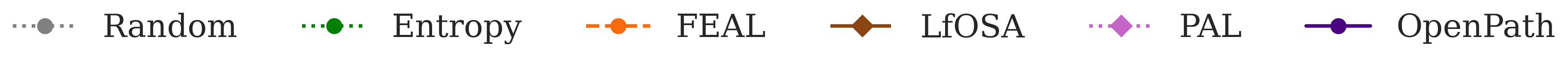}\\

    \includegraphics[width=0.24\textwidth]{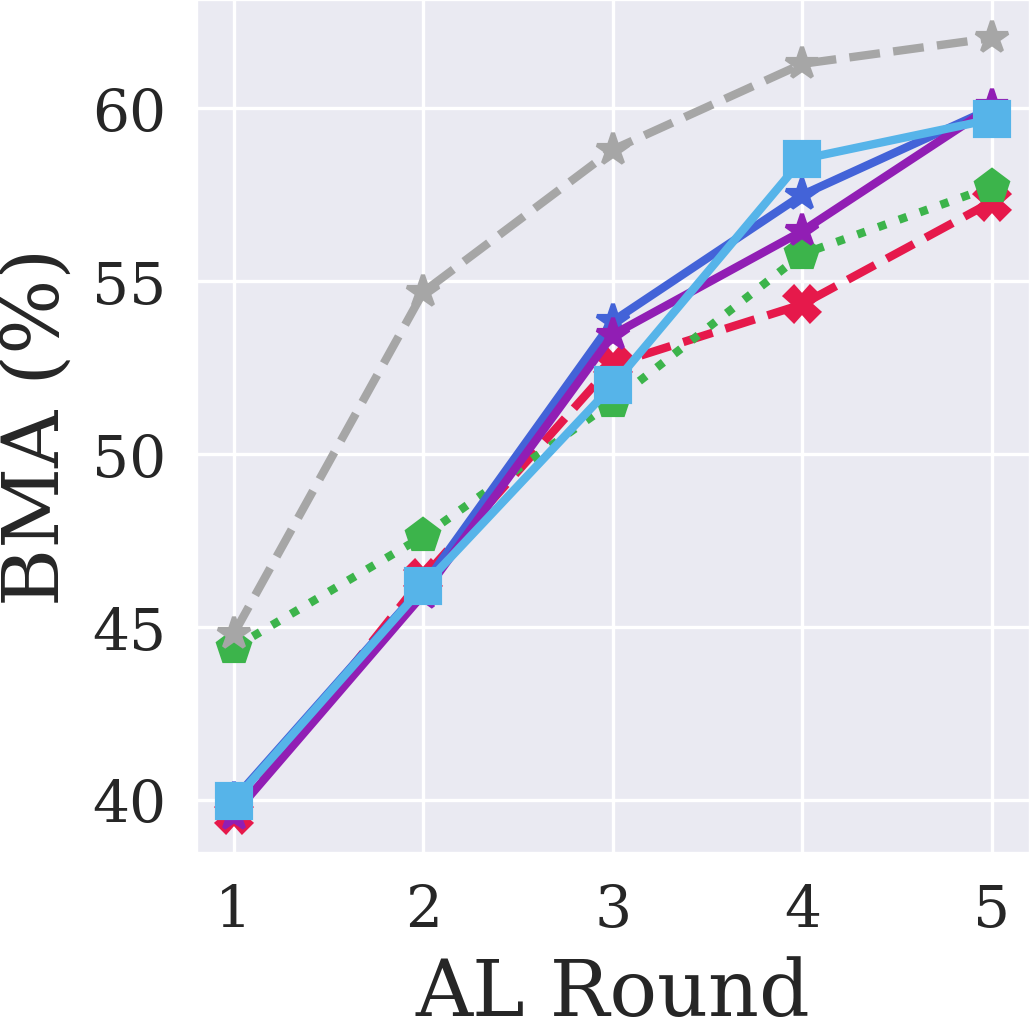}
    \includegraphics[width=0.24\textwidth]{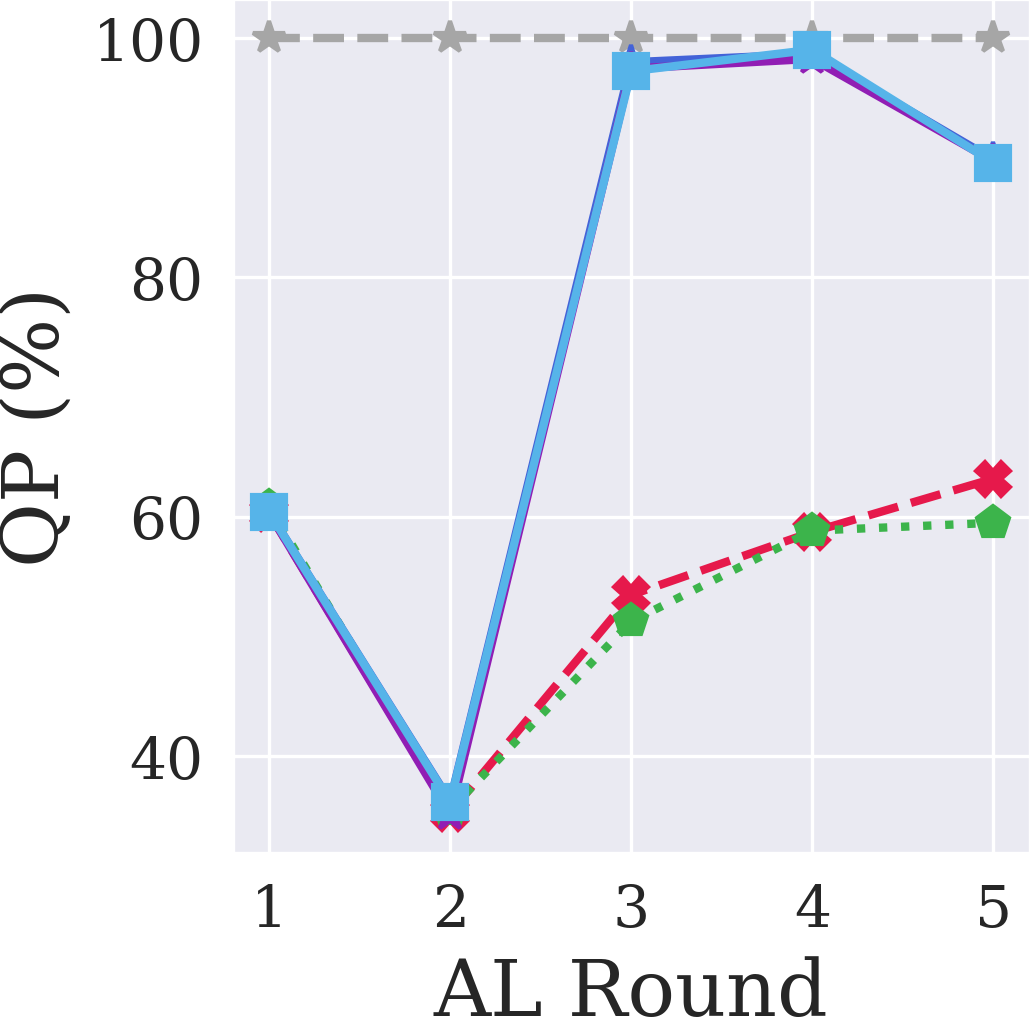}
    \includegraphics[width=0.24\textwidth]{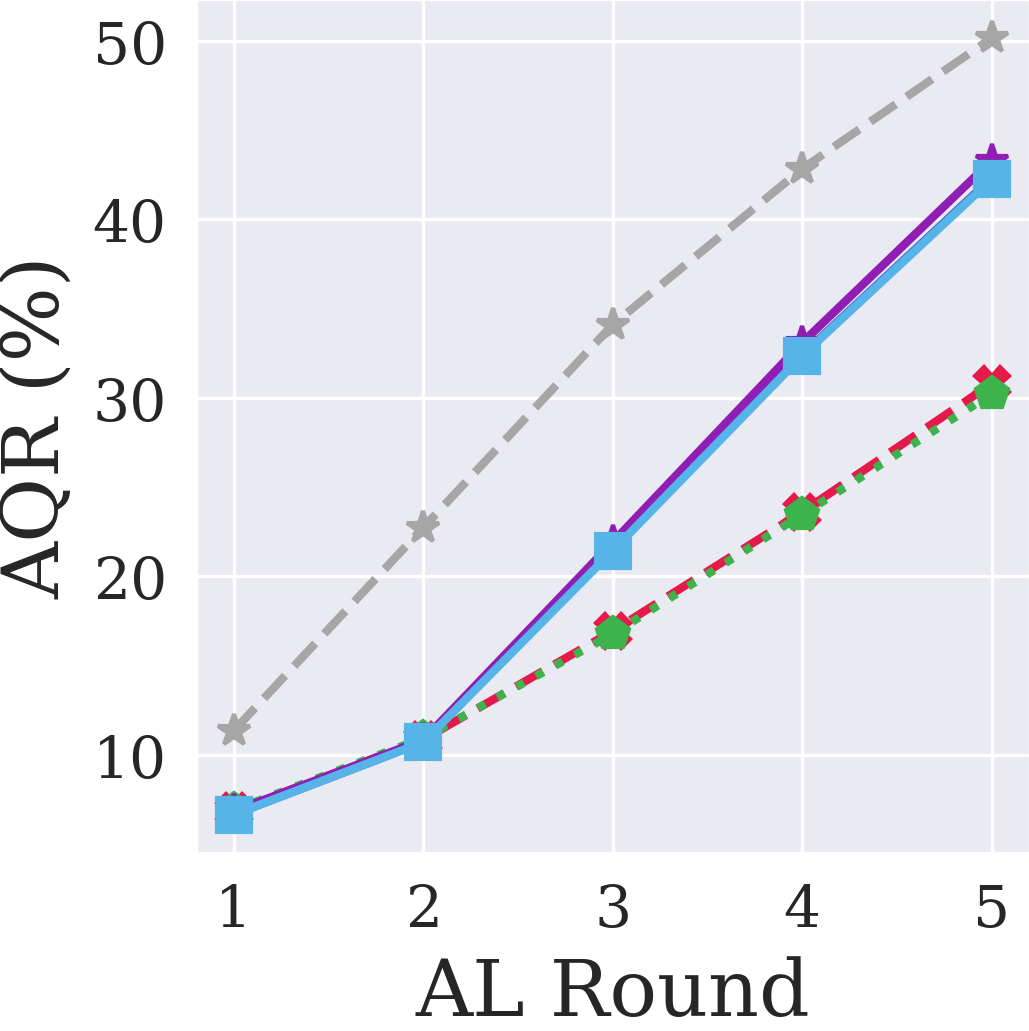}
    \includegraphics[width=0.24\textwidth]{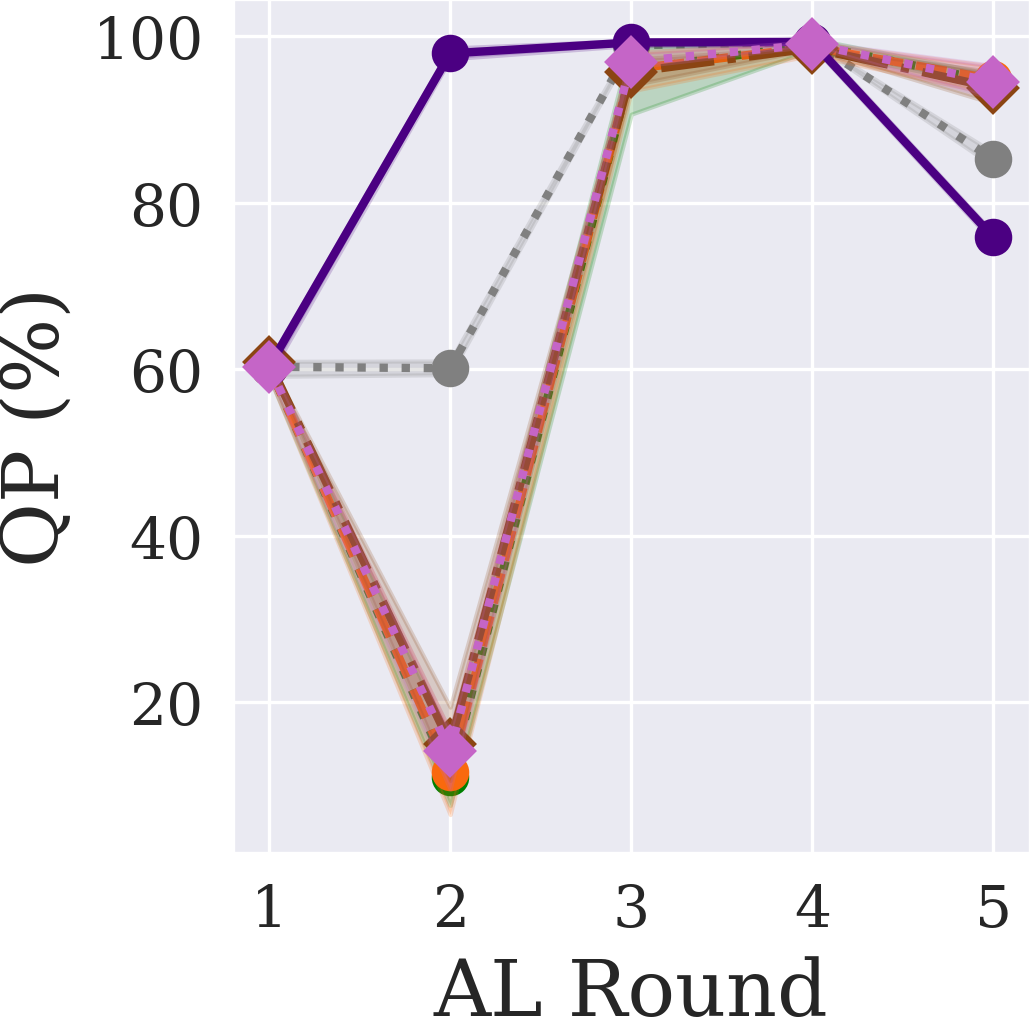}\\

    \includegraphics[width=0.24\textwidth]{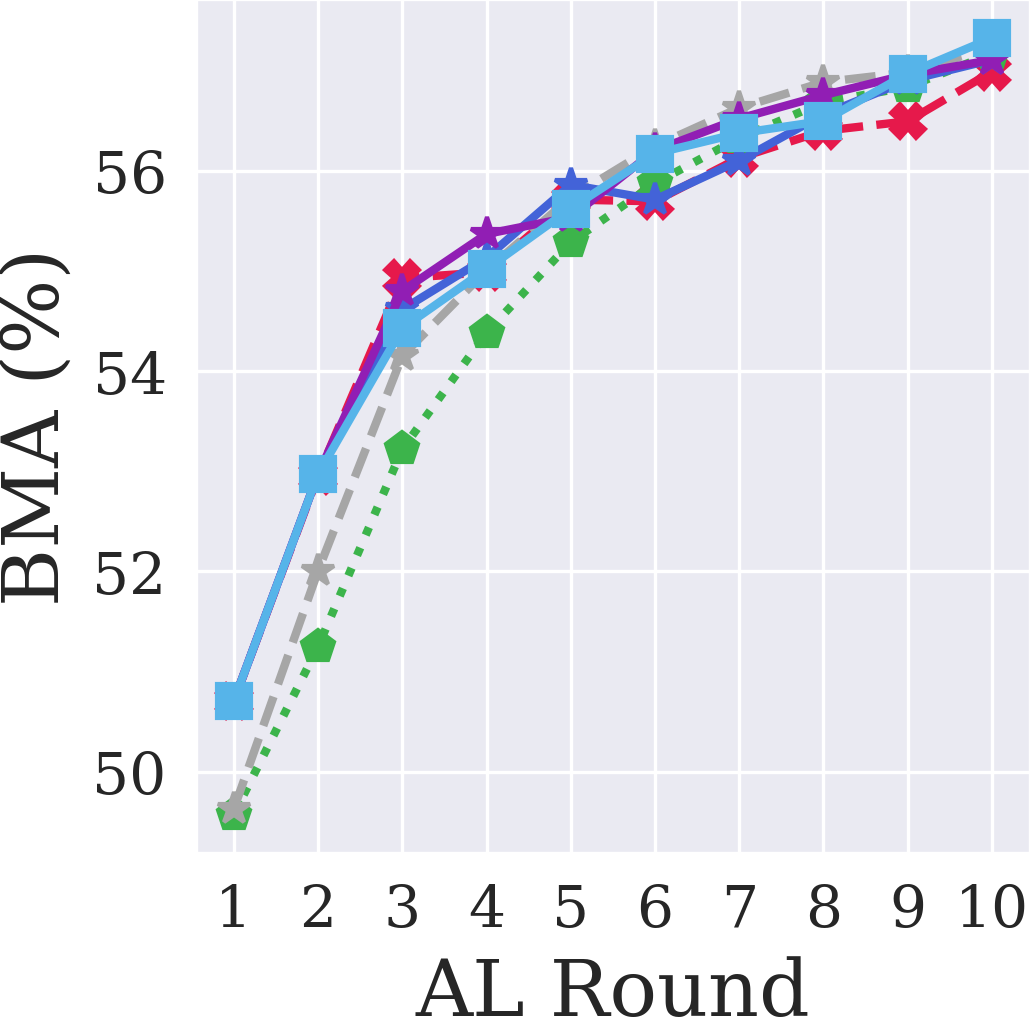}
    \includegraphics[width=0.24\textwidth]{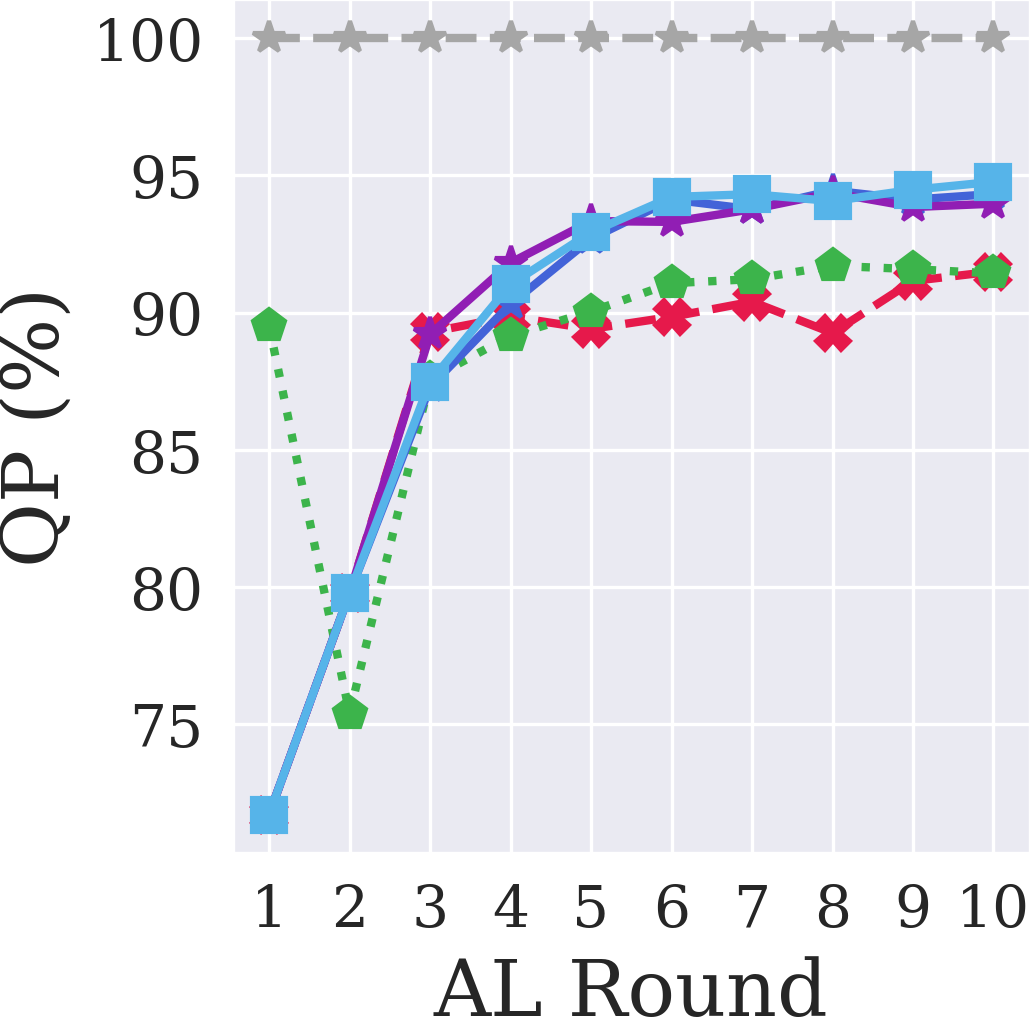}
    \includegraphics[width=0.24\textwidth]{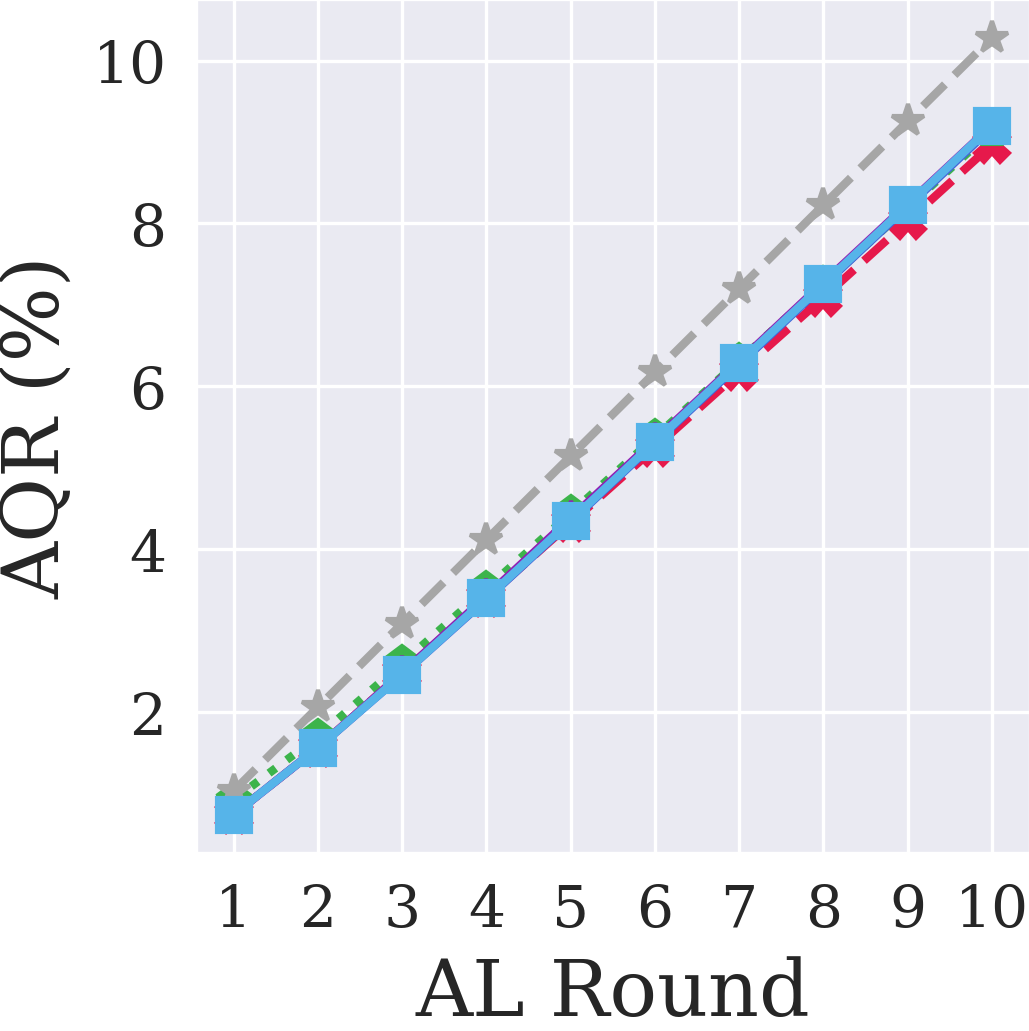}
    \includegraphics[width=0.24\textwidth]{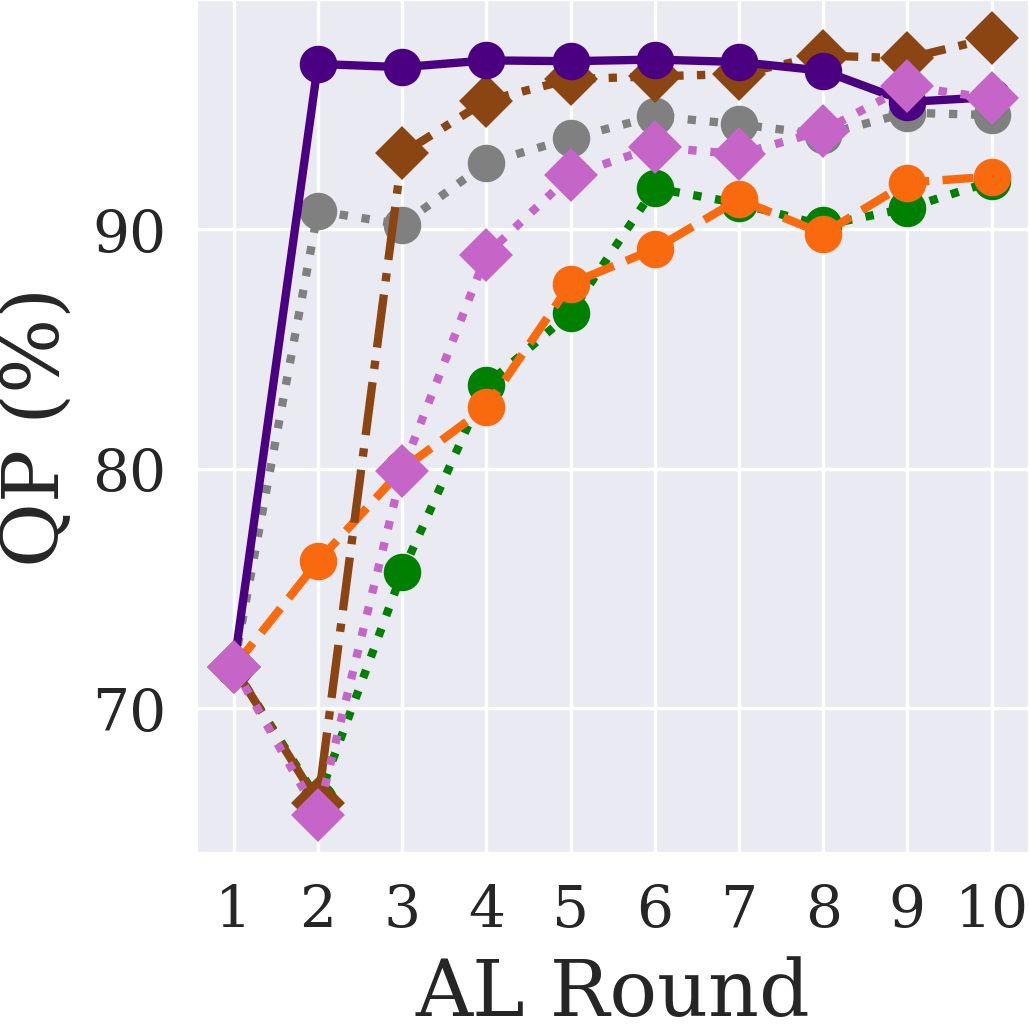}
    \caption{Average BMA, QP, and AQR across all OS-FAL methods for different VLM filter variants. Top row: FedISIC. Bottom row: FedEMBED.}
    \label{fig:vlm_variants_unified}
\end{figure*}

Table~\ref{tab:combined_purity_bma} compares OS-FAL methods under different VLM filtering strategies across FedISIC and FedEMBED. A key structural difference is that Coldstart operates on the full, unfiltered unlabeled pool, whereas VLM-based methods query from a pre-filtered ID-candidate pool; consequently, the two settings train on differently composed batches from R=1, which explains divergent early-round BMA and QP. Averaged across all strategies (Avg column), every PromptGate variant outperforms the Baseline in Purity on both benchmarks, with average gains of +22\% on FedISIC and +2--3\% on FedEMBED, and consistent BMA improvements of +1--3\%.
\textbf{FedISIC.} Without filtering, the Coldstart averages only 60.7\% Purity, while the static Baseline caps all strategies at $\leq$76\%, except OpenPath* (83.8\%), which benefits from stronger pretrained features. PromptGate variants achieve $>$97\% Purity uniformly (avg.\ 96.8\% for \emph{Local}), with \emph{Mixed} delivering the best overall BMA. As shown in Fig.~\ref{fig:vlm_variants_unified} (top row), VLM-based QP starts at ${\sim}60$\% at R=1, reflecting zero-shot gating prior to prompt adaptation, then temporarily dips at R=2 as the first CoOp update on a small 128-shot sample can shift the decision boundary sub-optimally; from R=3 onward accumulated labels stabilise the prompts and QP rises above 95\%. By contrast, Coldstart and Baseline QP steadily deteriorate as OOD samples accumulate. Under strong PromptGate filtering, closed-set strategies close the gap with open-set methods and in several cases surpass them, suggesting that a high-purity gate effectively converts the open-set problem into a closed-set one. Final BMA remains inherently dependent on the downstream acquisition strategy: information-dense methods like Entropy reach 66.1\% BMA under \emph{Mixed}, whereas Random achieves 64.4\%.

\textbf{FedEMBED.} Because organic OOD prevalence is low (4--13\% across scanners), the Coldstart already reaches 87.9\% average Purity and ${\sim}90$\% QP at R=1---most random queries are naturally ID. At R=2 QP drops to ${\sim}75$\% as uncertainty-driven strategies begin targeting ambiguous boundary samples that include subtle artefacts (e.g., circular markers); QP then recovers from R=3 onward as the easy ID pool is consumed and query composition stabilises. VLM-gated methods initially show slightly lower BMA (${\sim}$49\% vs.\ ${\sim}$51\% for Coldstart) because the filtered pool is smaller and somewhat less diverse; however, as prompt learning refines the boundary, this gap closes and PromptGate yields a net advantage. At the final round, \emph{Mixed} combined with OpenPath* reaches the highest Purity (96.0\%) and BMA (60.6\%), versus the Baseline's 95.0\%\,/\,60.5\%. Scanner-specific artefacts make fully local token adaptation the strongest per-client filter (avg.\ 90.7\% Purity), while the \emph{Mixed} variant's additional global context delivers the best peak when paired with a strong AL strategy.

\textbf{Adapter Strategy.} Ablating our approach (Table~\ref{tab:combined_purity_bma}) reveals that \textit{Local} strictly outperforms \textit{Global} and \textit{Mixed} aggregation. On FedISIC, the \textit{Local} adapter achieves the highest average Purity (96.8\%) and BMA (60.3\%). On FedEMBED, \textit{Local} maintains peak Purity (90.7\%) with a negligible (-0.2\%) BMA trade-off against \textit{Mixed}. This confirms that tailoring semantic alignment to client-specific artefacts yields a superior gatekeeper compared to generalised global prompts.

\begin{figure*}[t!]
    \centering
    \includegraphics[width=0.8\textwidth]{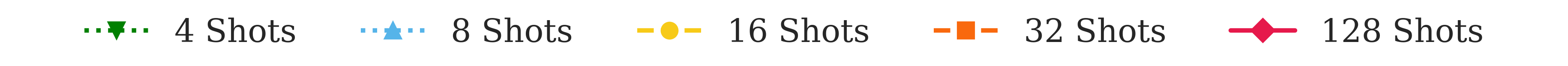} \\
        \includegraphics[width=0.32\textwidth]{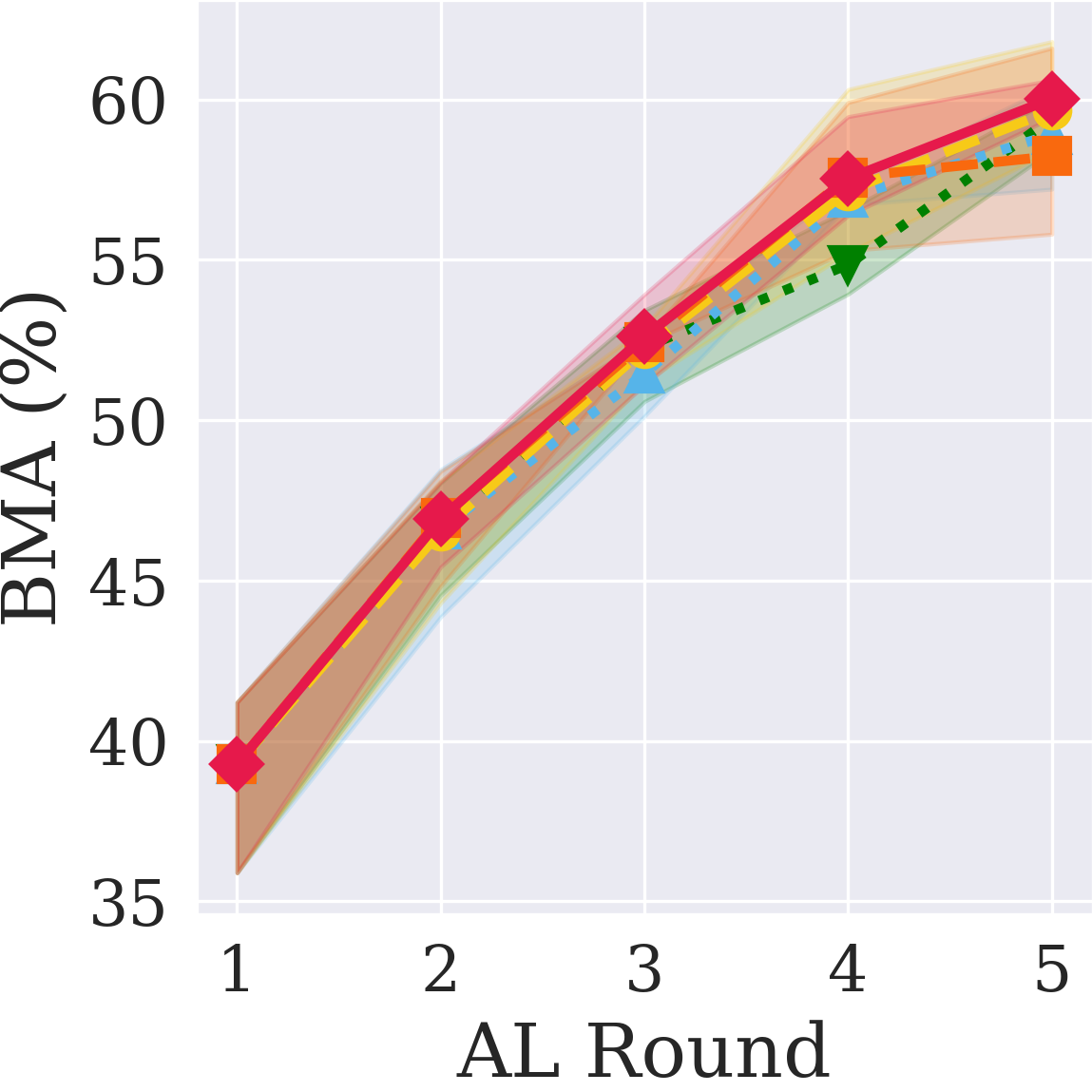}
        \includegraphics[width=0.32\textwidth]{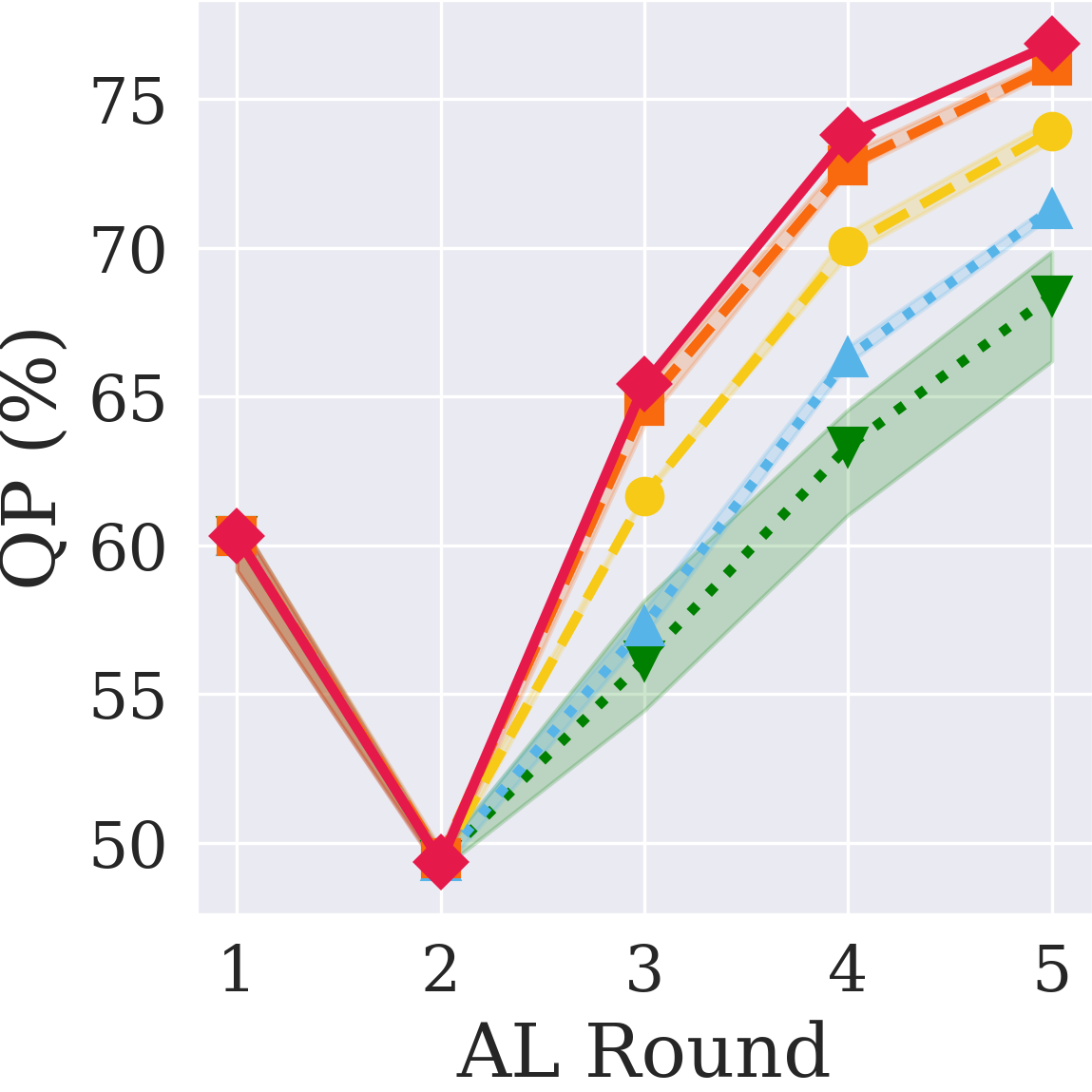}
        \includegraphics[width=0.32\textwidth]{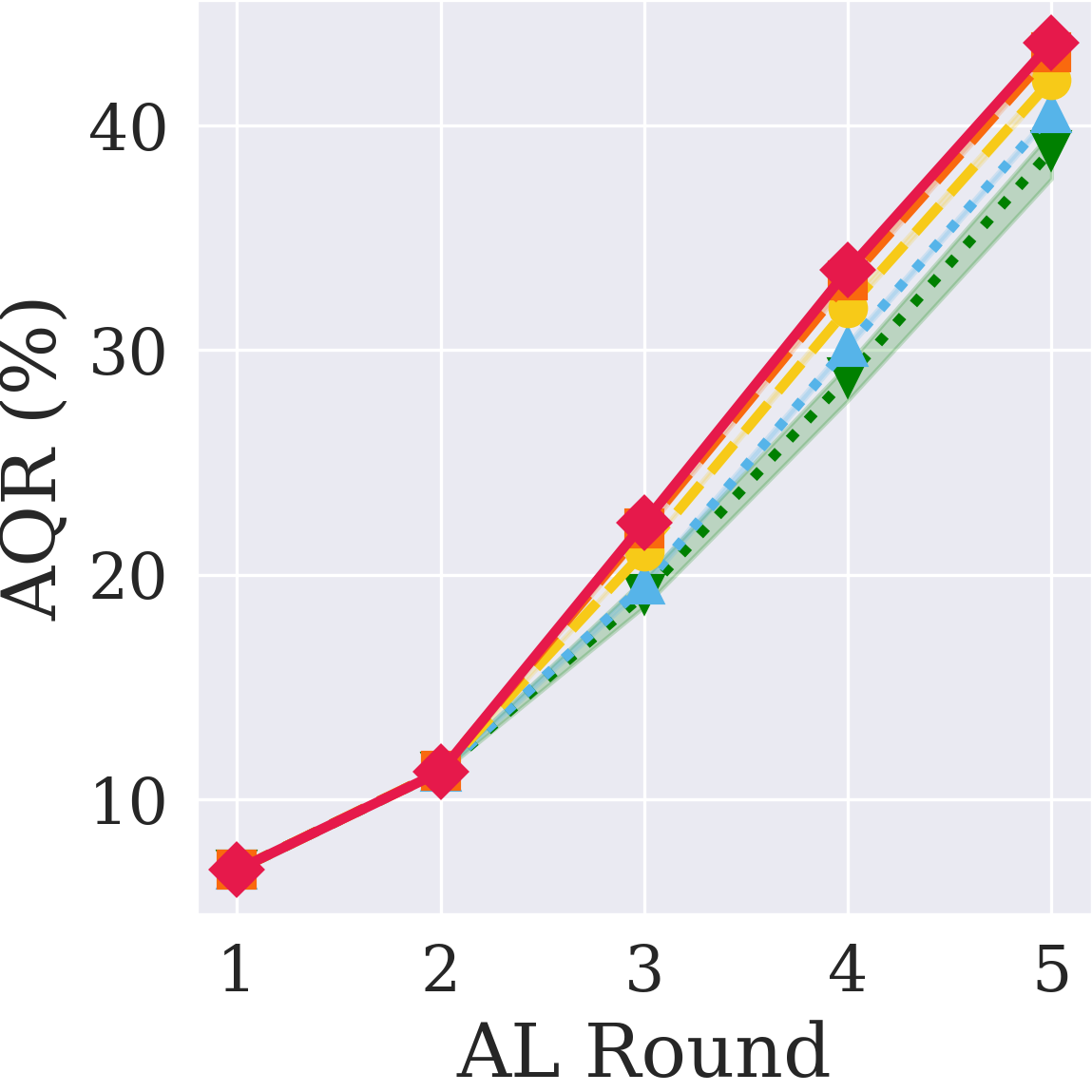} \\
        \includegraphics[width=\textwidth]{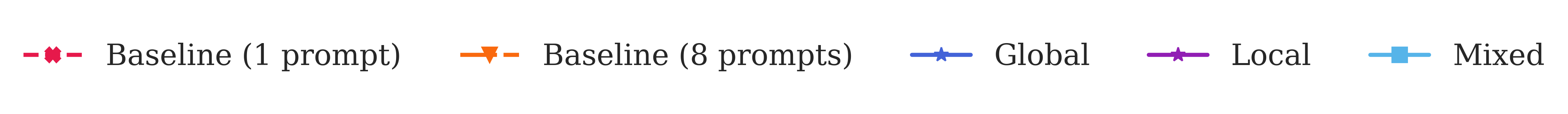} \\
        \includegraphics[width=0.32\textwidth]{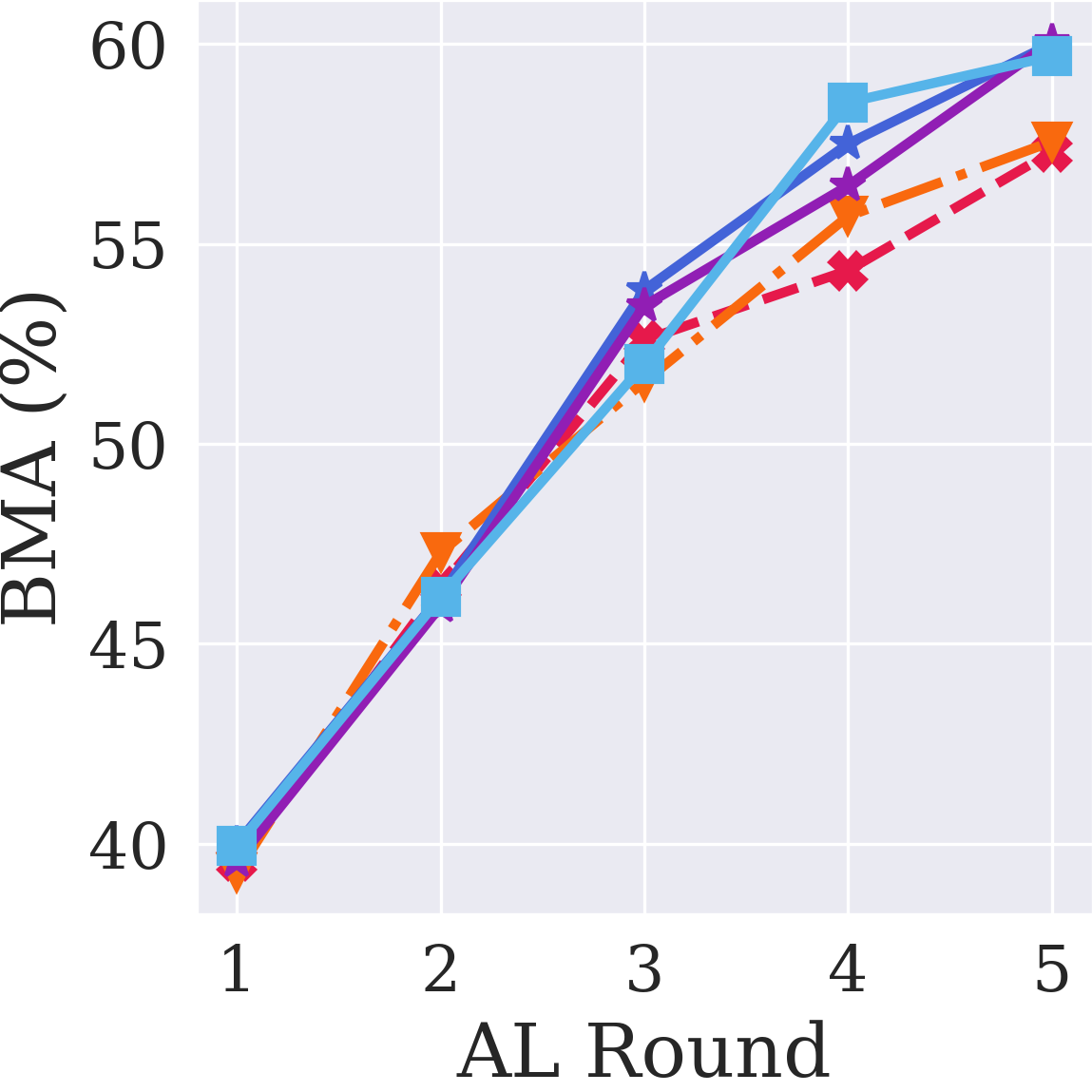}
        \includegraphics[width=0.32\textwidth]{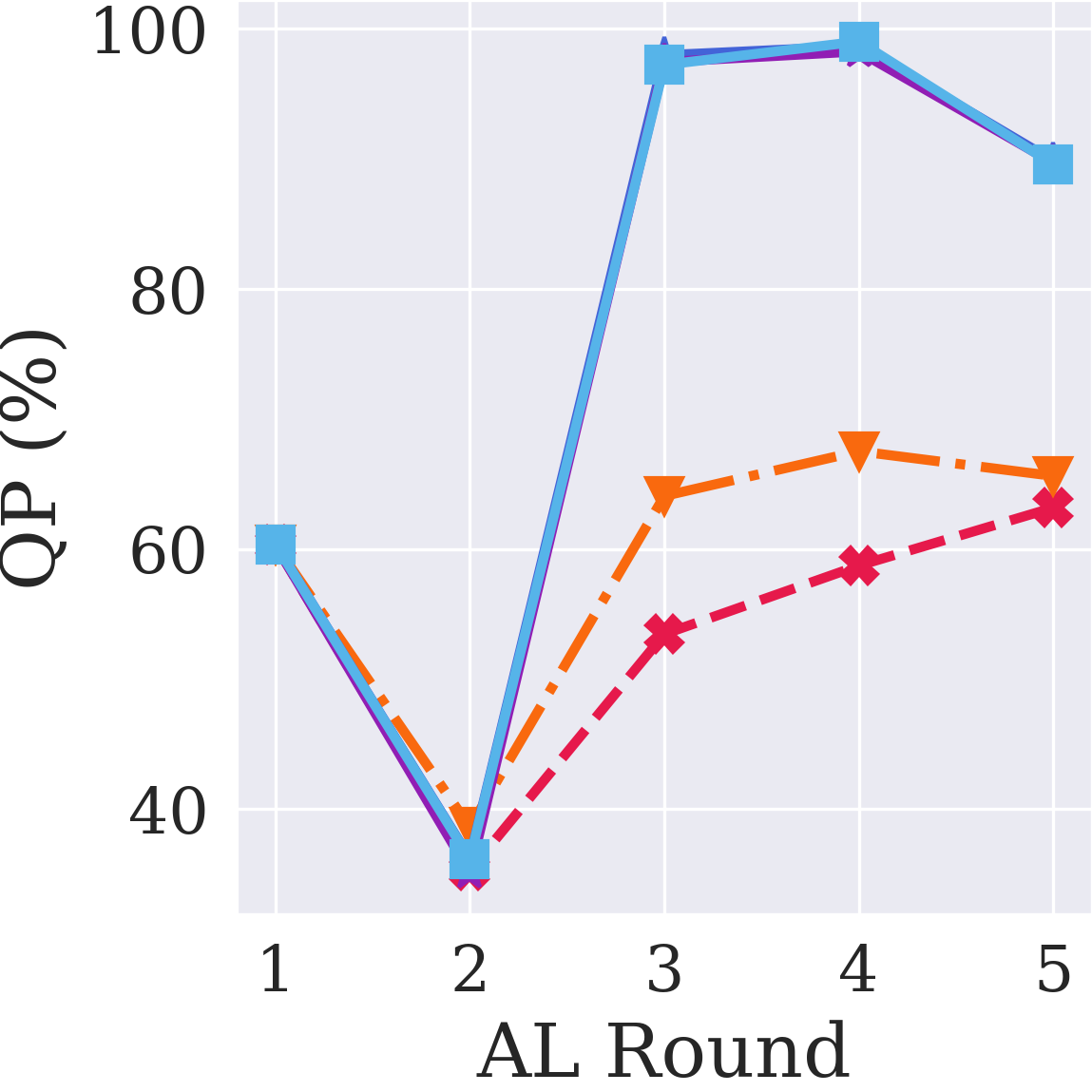}
        \includegraphics[width=0.32\textwidth]{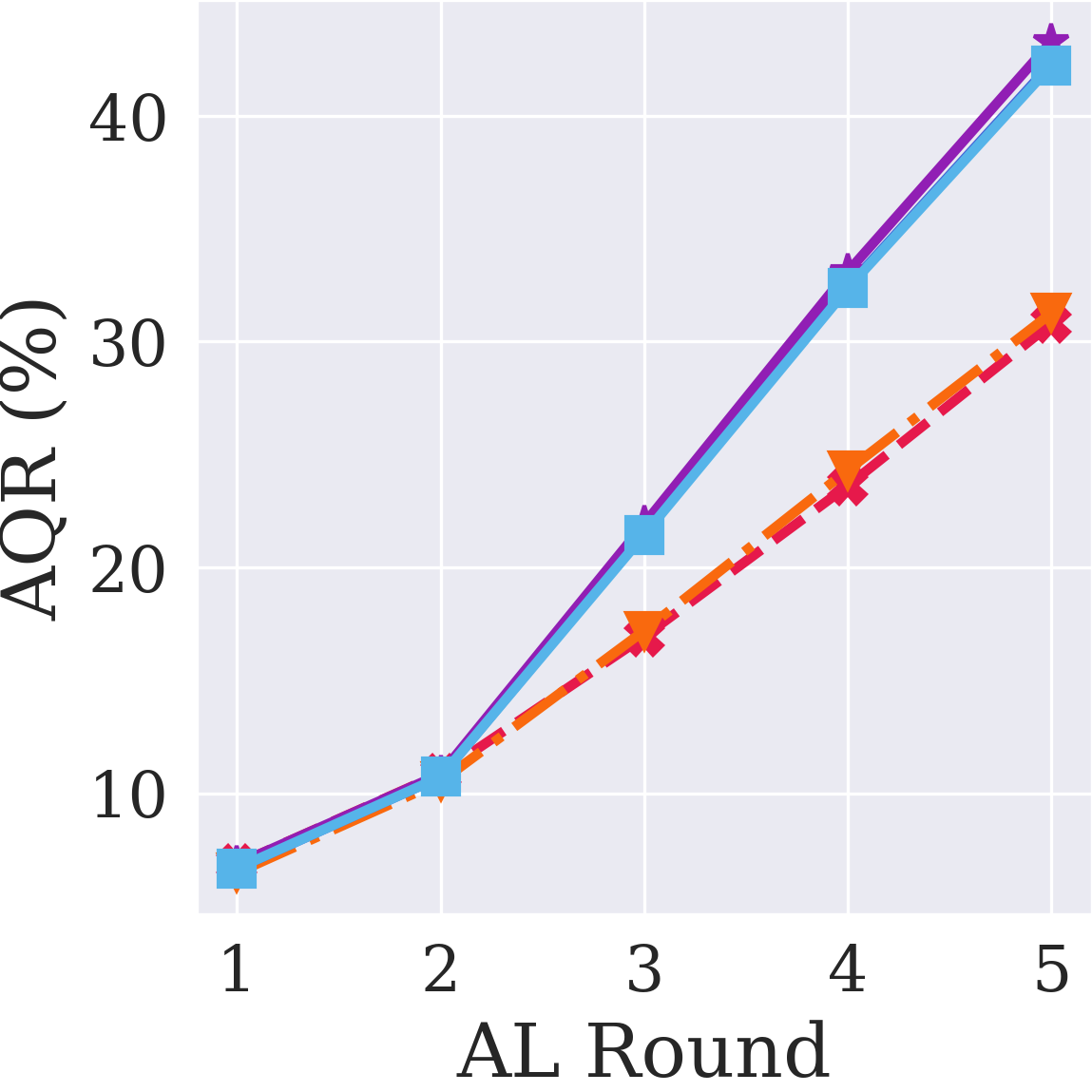}
    \caption{Average BMA, QP, and AQR across all AL methods for different VLM filter variants in FedISIC}
    \label{fig:shots_prompts_combined}
\end{figure*}

\begin{table}[h!]
 \centering
 \fontsize{8pt}{\baselineskip}\selectfont
        \caption{Last AL round VLM Test Set Performance on FedISIC (AL: Random).}         
        \label{tab:fedisic_test_vlm_performance}
        
        \setlength{\tabcolsep}{2.5pt} 
        \fontsize{8}{9.5}\selectfont 
  
        \begin{tabular}{clccccc}
        \toprule
        \textbf{Variant} & \textbf{Met. (\%)} & \textbf{C0} & \textbf{C1} & \textbf{C2} & \textbf{C3} & \textbf{Avg} \\
        \midrule
        \multirow{3}{*}{\textbf{Baseline}}
         & ID BMA   & 30.6 & 27.6 & 31.2 & 26.6 & 27.5 \\
         & Acc.     & 83.2 & 87.9 & 85.8 & 82.4 & 84.4 \\
         & OOD Recall   & 0.6 & 0.9 & 0.0 & 0.0 & 0.5 \\
        \midrule
        \multirow{3}{*}{\textbf{Ours (Mixed)}}
         & ID BMA   & \textbf{35.7} & \textbf{64.4} & 43.2 & 35.3 & \textbf{42.0} \\
         & Acc.     & 99.5 & 99.6 & \textbf{99.7} & 99.7 & \textbf{99.6} \\
         & OOD Recall   & 97.6 & 99.7 & \textbf{99.3} & 97.6 & 98.2 \\
         \midrule
        \multirow{3}{*}{\textbf{Ours (Global)}}
         & ID BMA   & 33.9 & 62.9 & \textbf{44.9} & 36.7 & 41.8 \\
         & Acc.     & \textbf{99.6} & 99.6 & \textbf{99.7} & \textbf{99.9} & \textbf{99.6} \\
         & OOD Recall   & 97.9 & 99.7 & \textbf{99.3} & 99.1 & 98.5 \\
        \midrule
        \multirow{3}{*}{\textbf{Ours (Local)}}
         & ID BMA   & 32.5 & 55.4 & 44.7 & \textbf{38.9} & 40.1 \\
         & Acc.     & \textbf{99.6} & 99.4 & \textbf{99.7} & 99.8 & \textbf{99.6} \\
         & OOD Recall   & \textbf{98.3} & \textbf{100.0} & 99.0 & \textbf{99.1} & \textbf{98.8} \\
        \bottomrule
        \end{tabular}%
\end{table}
\section{Discussion and Conclusion}
\label{sec:discussion}

\textbf{The paradox of expanding static OOD prompts.}
Expanding static VLM prompts with additional OOD descriptions (e.g., eight instead of one) seems a natural way to enhance OS detection. However, Fig.~\ref{fig:shots_prompts_combined} (bottom) shows a clear limitation: more static prompts yield only marginal gains, as one cannot anticipate the full diversity of clinical artifacts across clients. In contrast, dynamically adapting text embeddings to each local distribution provides a much more reliable gatekeeper. As shown in Fig.~\ref{fig:shots_prompts_combined} (top), even a small few-shot adaptation budget (4–128 shots) improves anomaly query recall by nearly 20\% while keeping downstream BMA stable (within $\pm1.5$\%). Thus, dedicating a small annotation fraction to prompt initialization is more effective than static prompt expansion, yielding robust and autonomous filtering in subsequent AL rounds.

\noindent\textbf{Limitations and future work.}
PromptGate's fine-grained ID classification remains limited (${\sim}$42\% BMA on FedISIC, Table~\ref{tab:fedisic_test_vlm_performance}), confirming that the VLM should act as an OOD gatekeeper rather than a standalone classifier. Early-round prompt adaptation can also temporarily reduce query precision (Section~\ref{sec:exp}), and the system inherits the VLM backbone's domain biases. Future work includes enforcing cross-modal agreement between the VLM and task model to stabilize early AL rounds, and selectively querying from the exploration pool. PromptGate itself introduces negligible overhead: only 16 prompt vectors (${\sim}$12K parameters) per client with a frozen backbone. Its privacy-preserving, plug-and-play design makes PromptGate readily deployable across heterogeneous hospital networks without centralized data curation or site-specific engineering.


\bibliographystyle{splncs04}
\bibliography{references}

\end{document}